%% file: main.tex
\documentclass[sigconf]{acmart}

\AtBeginDocument{%
  }

\copyrightyear{2023}
\acmYear{2023}
\setcopyright{acmlicensed}
\acmConference[MM '23]{Proceedings of the 31st ACM International Conference on Multimedia}{October 29--November 3, 2023}{Ottawa, ON, Canada.}
\acmBooktitle{Proceedings of the 31st ACM International Conference on Multimedia (MM '23), October 29--November 3, 2023, Ottawa, ON, Canada}
\acmPrice{15.00}
\acmISBN{979-8-4007-0108-5/23/10}
\acmDOI{10.1145/3581783.3612388}

\usepackage{multirow}
\usepackage{makecell}
\usepackage{bbding}
\usepackage{color}
\usepackage{hyperref}
\usepackage{colortbl}
\usepackage{balance}

\begin{document}
\begin{sloppypar}
\title[Enhancing Vision-Language Pre-Training with Jointly Learned Questioner and Dense Captioner]{\texorpdfstring{Enhancing Vision-Language Pre-Training with \\ Jointly Learned Questioner and Dense Captioner}{Enhancing Vision-Language Pre-Training with Jointly Learned Questioner and Dense Captioner}}

\author{Zikang Liu}
\email{liuzikang2023@ia.ac.cn}
\affiliation{%
  \institution{Institute of Automation, Chinese Academy of Sciences}
  \city{Beijing}
  \country{China}
}

\author{Sihan Chen}
\email{sihan.chen@nlpr.ia.ac.cn}
\affiliation{%
  \institution{Institute of Automation, Chinese Academy of Sciences}
  \city{Beijing}
  \country{China}
}

\author{Longteng Guo}
\email{longteng.guo@nlpr.ia.ac.cn}
\affiliation{%
  \institution{Institute of Automation, Chinese Academy of Sciences}
  \city{Beijing}
  \country{China}
}

\author{Handong Li}
\email{lihandong2023@ia.ac.cn}
\affiliation{%
  \institution{Institute of Automation, Chinese Academy of Sciences}
  \city{Beijing}
  \country{China}
}

\author{Xingjian He}
\email{xingjian.he@nlpr.ia.ac.cn}
\affiliation{%
  \institution{Institute of Automation, Chinese Academy of Sciences}
  \city{Beijing}
  \country{China}
}

\author{Jing Liu}
\email{jliu@nlpr.ia.ac.cn}
\authornote{Corresponding Author.}
\affiliation{%
  \institution{Institute of Automation, Chinese Academy of Sciences}
  \city{Beijing}
  \country{China}
}

\renewcommand{\shortauthors}{Zikang Liu et al.}

\begin{abstract}
Large pre-trained multimodal models have demonstrated significant success in a range of downstream tasks, including image captioning, image-text retrieval, visual question answering (VQA), etc. However, many of these methods rely on image-text pairs collected from the web as pre-training data and unfortunately overlook the need for fine-grained feature alignment between vision and language modalities, which requires detailed understanding of images and language expressions. While integrating VQA and dense captioning (DC) into pre-training can address this issue, acquiring image-question-answer as well as image-location-caption triplets is challenging and time-consuming. Additionally, publicly available datasets for VQA and dense captioning are typically limited in scale due to manual data collection and labeling efforts. In this paper, we propose a novel method called \textbf{J}oint Q\textbf{A} and \textbf{D}C G\textbf{E}neration (JADE), which utilizes a pre-trained multimodal model and easily-crawled image-text pairs to automatically generate and filter large-scale VQA and dense captioning datasets. We apply this method to the Conceptual Caption (CC3M) dataset to generate a new dataset called CC3M-QA-DC. Experiments show that when used for pre-training in a multi-task manner, CC3M-QA-DC can improve the performance with various backbones on various downstream tasks. Furthermore, our generated CC3M-QA-DC can be combined with larger image-text datasets (e.g., CC15M) and achieve competitive results compared with models using much more data. Code and dataset are available at \href{https://github.com/johncaged/OPT\_Questioner}{https://github.com/johncaged/OPT\_Questioner}.
\end{abstract}




\maketitle

\section{Introduction}

\begin{figure}[!htbp]
  \centering
  \includegraphics[width=\linewidth]{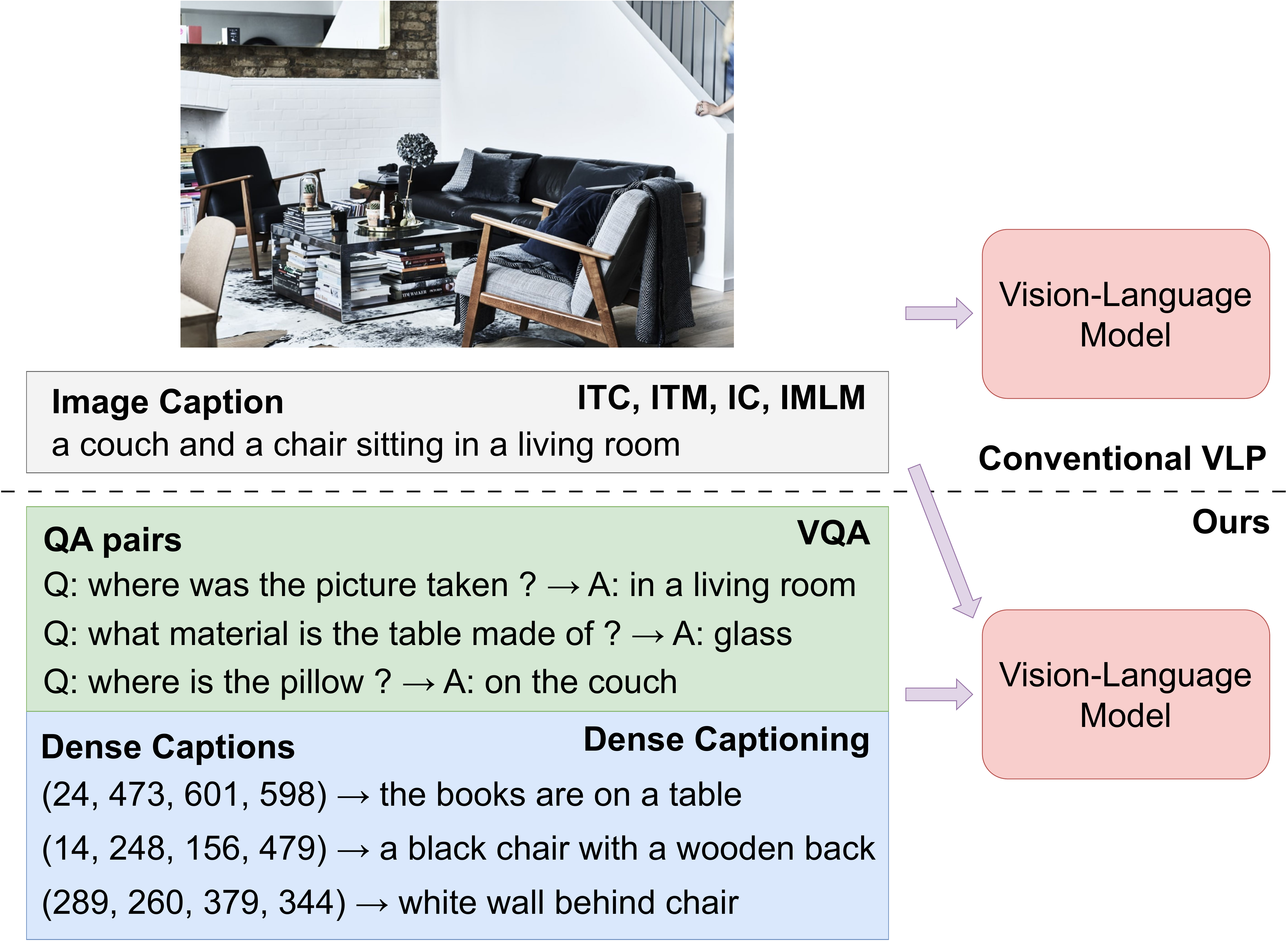}
  \caption{Conventional vision-language pre-training (VLP) paradigm (top) using the original CC3M dataset (gray) and our proposed VLP paradigm (bottom) additionally using our generated QA (green) and DC (blue) data.}
  \label{fig:intro}
\end{figure}

Vision-language pre-training has made remarkable progress in recent years, with new large pre-trained models emerging constantly. As a common practice, the vast majority of these models utilize image-text datasets crawled from the web for pre-training purposes. The objective is to enhance the model's understanding of the vision and language modalities. Based on these datasets, a wide range of pre-training tasks have been proposed, including Image-Text Matching (ITM) \cite{li2021align,chen2020uniter,huang2020pixel,lu2019vilbert}, Image-Text Contrastive Learning (ITC) \cite{radford2021learning,pham2021combined,zhai2022lit,yuan2021florence}, Image Captioning \cite{cho2021unifying,wang2022ofa}, Image-Conditioned Masked Language Modeling (IMLM) \cite{li2021align,yu2022coca,wang2022image}, among others. These methods have significantly improved large pre-trained models in multimodal understanding and generation.

However, we argue that the above methods focus on coarse-grained learning and lack fine-grained alignment between images and text, which involves precise understanding of vision, language and the detailed correlation between them (e.g., answering questions or generating captions about local regions in images), and can further enhance the capabilities of pre-trained models. We assume Visual Question Answering (VQA) and Dense Captioning (DC) are two ideal training tasks that can help address this issue.

On the one hand, VQA requires exact understanding of both global and local details of images and aligns images with question queries in a fine-grained manner to generate correct answers based on the semantics of both image and text. On the other hand, DC, which means generating captions given bounding boxes, involves learning the location and detailed semantics of images, thus enriching text representation of the vision modality. Therefore, these two tasks are essentially complementary to the existing image-text pre-training paradigm, and their integration should significantly improve the performance of pre-trained models in multimodal understanding and generation. Moreover, due to the development of generalist models \cite{li2022uni} and chatbots \cite{ouyang2022training}, zero-shot VQA and DC are receiving increasingly more attention, highlighting the needs to include these tasks in the pre-training process.

Some methods \cite{wang2022ofa} have incorporated human-annotated VQA or DC datasets in their pre-training. However, unlike image-text pairs that can be obtained from the internet at a relatively low cost, image-question-answer and image-location-caption triplets for VQA and DC are much more challenging to collect. These datasets are often labeled manually, which can produce high-quality and carefully-examined data. Nevertheless, manual labeling is a time-consuming and laborious process. Due to this issue, publicly available VQA and DC datasets are mostly limited in scale, making them sub-optimal choices for large-scale pre-training.

To address this challenge, we propose a novel method called Joint QA and DC Generation (JADE), which leverages a large pre-trained multimodal model and public image-text datasets to automatically generate and filter VQA and DC data. By doing so, we can obtain a sufficient amount of high-quality data for pre-training. Figure \ref{fig:intro} illustrates the conventional vision-language pre-training (VLP) paradigm using the original CC3M \cite{sharma2018conceptual} dataset and our proposed VLP paradigm which adopts additional QA (green) and DC (blue) data generated by our JADE method. All these data can be jointly utilized for multi-task pre-training, leading to improved performance.

Our JADE method can be summarized in three steps. Firstly, the pre-trained model is fine-tuned on manually annotated VQA and DC datasets to learn dense captioning and conditional question-answer pair generation (referred to as the generator model). Dense captions are generated based on images and bounding box coordinates, while the conditional question-answer pairs are produced in a sequence-to-sequence manner, given images, question types, bounding box coordinates, and corresponding dense captions. The use of the above prompts makes the generation process both content and question type controllable. Secondly, to further improve data quality and reduce generation bias, we also fine-tune the same pre-trained model on VQA datasets to create a common VQA model (referred to as the filter model). During inference, the filter model predicts answers based on images and generated questions. Finally, the predicted answers from the filter model and the generated answers from the generator model are compared, and unmatched QA pairs are discarded. This step further improves the quality of the generated VQA data.

We apply our JADE method to build a new VQA and DC dataset derived from the CC3M dataset, which we name CC3M-QA-DC. We use the CC3M-QA-DC dataset in vision-language pre-training with various vision and language backbones in a multi-task manner. Experiments demonstrate that our CC3M-QA-DC dataset efficiently helps multimodal models learn image-text alignment and cross-modality fusion, leading to improved performance on a wide range of downstream tasks, such as image captioning, image-text retrieval and VQA. Our method has shown great promise in AI Generated Data (AIGD), which contributes to new data acquisition methods and lowers data collection costs. Furthermore, we find that our large-scale CC3M-QA-DC dataset, employed in pre-training, can enhance zero-shot performance on the aforementioned downstream tasks. This finding highlights the potential for training generalist models and developing strong multimodal chatbots.

Overall, our contribution can be summarized as follows:
\begin{enumerate}
    \item We propose a novel method called Joint QA and DC Generation (JADE) to generate and filter QA and DC data in an end-to-end manner, by leveraging a large pre-trained multimodal model and public image-text datasets. This method enables the efficient collection of high-quality data for pre-training, even at a large scale.
    \item We apply JADE to produce QA and DC data derived from the CC3M dataset, resulting in a large-scale VQA and DC dataset called CC3M-QA-DC, which covers a wide range of question types and is beneficial for training vision-language models.
    \item Experiments demonstrate that when used in pre-training, our CC3M-QA-DC dataset can improve multimodal models' zero-shot as well as fine-tuning performance on downstream tasks. These findings highlight the potential of AI Generated Data (AIGD) and training generalist multimodal models.
\end{enumerate}

\section{Related Work}

\subsection{Vision-Language Pre-Training}

In recent years, there has been a growing interest in developing cross-modal pre-training models that demonstrate strong capabilities and generalizability across various domains. Among these, the vision-language pre-training remains the predominant focus, giving rise to many representative works \cite{li2021align,wang2022image,yu2022coca,li2022blip,wang2022ofa,li2023blip,radford2021learning,alayrac2022flamingo}. In these existing approaches, the primary pre-training tasks include Image-Text Contrastive Learning (ITC), Image-Text Matching (ITM), Image Captioning (IC), and Image-Conditioned Masked Language Modeling (IMLM). Specifically, ITC \cite{radford2021learning,yaofilip,li2021align} employs contrastive learning to learn a shared image-text embedding space, which gathers image features to the corresponding text features and learns the alignment between vision and language modalities. ITM \cite{li2021align,li2022blip,bao2022vlmo} uses a classification head to determine if an image and text match, thereby extracting cross-modal information. As for generation tasks, a common practice is learning to generate text sequence in an auto-regressive manner. These existing tasks have been proven complementary in many studies and exhibit strong understanding and generalization capabilities for a wide range of downstream tasks.

However, for certain downstream tasks that require more fine-grained discriminative and generative abilities, such as Visual Question Answering (VQA) and Dense Captioning (DC), existing pre-training tasks based on image-text pairs tend to focus more on global modeling of the two modalities and lose local information. Fine-tuning on fine-grained datasets subsequently becomes challenging due to the inconsistencies between the two stages. We believe that introducing tasks that aid fine-grained semantic understanding during the pre-training phase can address this issue. There already exist works that incorporate VQA into vision-language pre-training, such as OFA \cite{wang2022ofa}, LXMERT \cite{tan2019lxmert} and VL-BART \cite{cho2021unifying}. However, these methods utilize limited human-annotated VQA datasets, which are sub-optimal for larger-scale vision-language pre-training. Moreover, most existing cross-modal pre-training models lack zero-shot chat capabilities, primarily due to the absence of relevant pre-training tasks during the training process. Our method can effectively resolve this problem, providing basic conversational abilities that contribute to the future development of chatbots.

\begin{figure*}[!htbp]
  \centering
  \includegraphics[width=\linewidth]{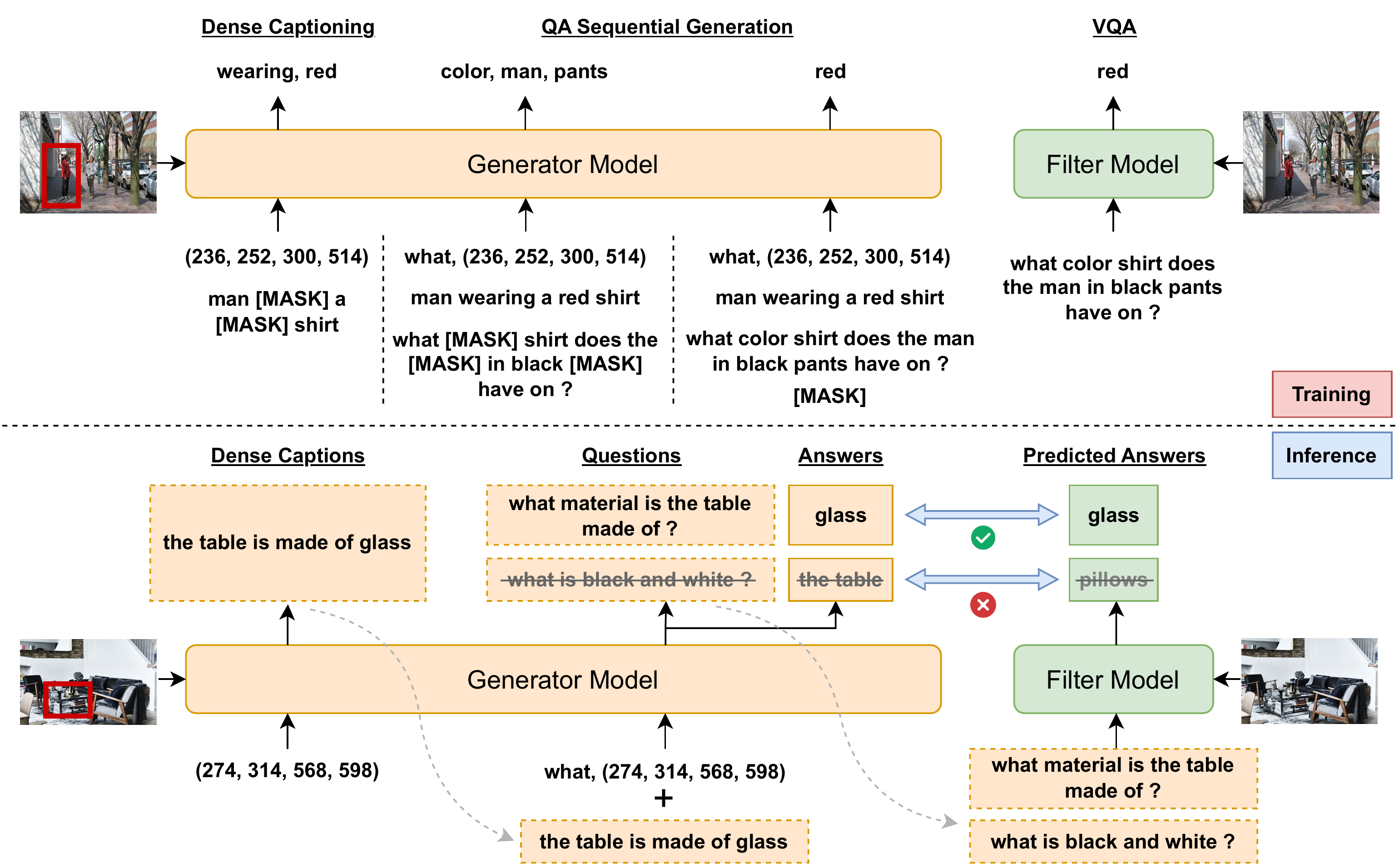}
  \caption{Overall pipeline of our Joint QA and DC Generation (JADE) method. We train a generator model and a filter model. During training stage (top), the generator model is fine-tuned in a multi-task manner, while the filter model is trained as a common VQA model. During inference stage (bottom), given corresponding prompts, the generator model first produces dense captions according to the bounding boxes and then generates QA pairs one at a time, while the filter model predicts answers given the generated questions. Matched QA pairs and dense captions are kept.}
  \label{fig:pipeline}
\end{figure*}

\subsection{Pre-Training Data Generation}

For a considerable duration, researchers have designed a range of generative downstream tasks, such as Image Captioning (IC) \cite{vinyals2015show}, Image Dense Captioning (DC) \cite{johnson2016densecap}, Visual Question Generation (VQG) \cite{mostafazadeh2016generating}, etc. The Image Captioning task aims to generate captions for images, and due to the availability of large datasets \cite{chen2015microsoft,sharma2018conceptual,changpinyo2021conceptual,jia2021scaling}, it has been widely applied in visual-language pre-training tasks. A similar task, Dense Captioning, requires generating more detailed natural language descriptions based on local image regions. In the past few years, a variety of works \cite{li2019learning,johnson2016densecap,shao2022region,yang2017dense,yin2019context,wu2022grit} have combined Dense Captioning with object detection tasks, with the primary dataset being Visual Genome (VG) \cite{johnson2016densecap}. Visual Question Generation (VQG) aims to generate visually-related questions based on an image or video. Initially a standalone task, it has gradually evolved into a means of data augmentation and evaluation to help improve Visual Question Answering (VQA) tasks \cite{kafle2017data,li2018visual,shah2019cycle,xu2020radial,kil2021discovering,akula2021crossvqa}.

Recently, the advancement of large pre-trained models has rendered the above mentioned generative downstream tasks as a feasible approach for generating data that can be utilized for pre-training purposes. One of the representitive methods, BLIP \cite{li2022blip}, proposed to use large pre-trained models to generate cleaner image captions than the weakly image-correlated alt-text collected from the web. In the VQG task, VQ$^{2}$A \cite{changpinyo2022all} selects keywords from the captions of existing image-text pairs as answers, then generates questions based on the answers and the original captions, automatically creating image-question-answer triplets. However, their approach utilizes weakly-correlated image-text data, which is purely linguistic and thus incapable of addressing detailed image-related questions.

Therefore, we propose our JADE method to automatically generate more abundant VQA and dense captioning data in an end-to-end manner, by leveraging a large pre-trained multimodal model and web-crawled image-text pairs. Except for VQA data, our method is capable of producing dense captioning data, which has been proved to be advantageous for multimodal understanding and generation. Moreover, unlike VQ$^{2}$A, we employ our generated dataset for multi-task pre-training and investigate diverse experiment settings, providing valuable insights into the incorporation of additional pre-training tasks and the development of generalist models as well as chatbots.

\section{Method}

Large pre-trained multimodal models have shown impressive capabilities in cross-modality understanding and generation. Building on this foundation, we propose to leverage these capabilities to address the challenge of collecting high-quality VQA and DC datasets at scale. Specifically, we utilize the VALOR \cite{chen2023valor} model as the fundamental model for our JADE method. The overall pipeline of our method is illustrated in Figure \ref{fig:pipeline}.

\subsection{QA and DC Generator Model}

The QA and DC Generator Model is fine-tuned in a multi-task way, incorporating two tasks named dense captioning and QA sequential generation.

\subsubsection{Dense Captioning}

In the dense captioning task, the model takes images and bounding box coordinates as input and learns to generate dense captions based on the given coordinates.

Given a single set of bounding box coordinates $ (x1, y1, x2, y2) $, which represents the horizontal and vertical coordinates in the upper left and lower right corner of a target bounding box in an input image, we first apply target embedding to the output feature map from the first vision layer. Let $ \boldsymbol{e_{t}} $ be the target embedding vector and $ \boldsymbol{e_{\bar{t}}} $ be the non-target embedding vector, the process can be denoted as:
\begin{equation}
    \boldsymbol{E_{i, j}} = \begin{cases}
        \boldsymbol{e_{t}}, \ if \ x1 \leq i \leq x2 \ and \ y1 \leq j \leq y2 \\
        \boldsymbol{e_{\bar{t}}}, \ otherwise
    \end{cases}
\end{equation}
\begin{equation}
    \boldsymbol{F}_{vision} = \boldsymbol{f}_{vision} + Conv(\boldsymbol{E})
\end{equation}
where $ \boldsymbol{E} \in \mathbb{R}^{3 \times H \times W} $, $ \boldsymbol{f}_{vision} \in \mathbb{R}^{C \times \frac{H}{p} \times \frac{W}{p}} $, $ \boldsymbol{F}_{vision} \in \mathbb{R}^{C \times \frac{H}{p} \times \frac{W}{p}} $ are the embedding map, output feature map from the first vision layer and finally embedded feature map respectively, $ C $ denotes the hidden dim and $ p $ denotes the patch size. 

To further represent the target area features, we use positional tokens as the prompt input to the language backbone, which are in the form of \textbf{\textit{[BOS] [LOC] [EOS]}}, where \textbf{\textit{[BOS]}} and \textbf{\textit{[EOS]}} denote beginning of sentence and end of sentence tokens, and \textbf{\textit{[LOC]}} denotes bounding box coordinate tokens. The model learns to generate dense captions with respect to the bounding boxes using causal masked language modeling, which can be written as:
\begin{equation}
    \mathcal{L}_{MLM}(\theta, w) = -E_{(w,v)\sim{D}}logP_{\theta}(w_{m} | w_{<m},v)
\end{equation}
where $ w $ denotes the target text sequence, $ v $ denotes the vision features from the vision backbone, $ m $ denotes the current index of the output target token and $ D $ denotes the dataset.

Therefore, the loss function of the dense captioning task can be described as:
\begin{equation}
    \mathcal{L}_{DC} = \mathcal{L}_{MLM}(\theta, w_{DC})
\end{equation}

The dense captioning task can help our model learn the correlation between local regions and output contents. Additionally, the generated dense captions can also be utilized to generate QA pairs and can be adopted in pre-training tasks.

\subsubsection{QA Sequential Generation}

In the QA sequential generation task, the model learns to generate condition-guided QA pairs in a sequence-to-sequence manner.

Given a single set of bounding box coordinates $ (x1, y1, x2, y2) $, same as the dense captioning task, we apply target embedding and positional token prompt in the forward pass. Furthermore, to make the generation process more controllable in both question type and content, we add question types (\textit{What}, \textit{How}, \textit{Where}, \textit{Binary}, etc.) and dense captions as part of the prompt into the language backbone. In general, the prompt input in the QA sequential generation task can be described in the form of \textbf{\textit{[BOS] [question type] [Task SEP] [LOC] [Task SEP] [dense caption] [EOS]}}, where \textbf{\textit{[Task SEP]}} is the special token to separate task prompts.

The target QA output sequence is in the form of \textbf{\textit{[BOS] [question] [QA SEP] [answer] [EOS]}}, where \textbf{\textit{[QA SEP]}} denotes the special token to separate question and answer sequences. During training, we randomly mask the question sequences with probability $ p $ and apply full mask to the answer sequences to get more accurate answers during inference. The QA sequential generation task is learned using a two-task approach. In the first task, the target sequence is the randomly masked question to promote question generation, while in the second task, the target sequence is the full question and the fully masked answer, to keep consistent with the inference stage, during which all question tokens are available when generating answers. The two loss functions of the two tasks can be denoted as:
\begin{equation}
    \mathcal{L}_{Q} = \mathcal{L}_{MLM}(\theta, w_{Q_{masked}})
\end{equation}
\begin{equation}
    \mathcal{L}_{A} = \mathcal{L}_{MLM}(\theta, Concat(w_{Q}, w_{A_{masked}}))
\end{equation}

The loss function of the QA and DC generator model is the sum of the above three loss functions:
\begin{equation}
    \mathcal{L}_{Gen} = \mathcal{L}_{DC} + \mathcal{L}_{Q} + \mathcal{L}_{A}
\end{equation}

During inference stage, our method utilizes top-K sampling to generate dense captions and a two-stage sampling strategy for QA generation. Specifically, in the two-stage sampling, the model first uses top-K sampling to obtain abundant questions, and then adopts greedy sampling to get relatively accurate answers. This strategy can be easily applied in a batch, by detecting \textbf{\textit{[QA SEP]}} tokens among samples.

\subsection{QA Filter Model}

In order to discard inaccurate, vague, or meaningless QA pairs generated by the QA and DC generator model, we employ a QA filter model that is fine-tuned using the same dataset and initial pre-trained weight as the generator model. The QA filter model is trained as a common VQA model, taking images and questions as input and the corresponding answers as output targets. During inference stage, the QA filter model predicts answers given the images and generated questions. We then discard QA pairs whose generated answers from the generator model and predicted answers from the filter model do not match exactly. This means that these QA pairs may be of low quality and could have a detrimental effect on training. During inference, we use beam-search sampling with a beam size of $ b $ for answer prediction.

\subsection{CC3M-QA-DC Dataset}

We utilize our JADE method to generate QA and DC data derived from the CC3M \cite{sharma2018conceptual} dataset, and create a new dataset named CC3M-QA-DC. We extracted object bounding box coordinates in advance using a bottom-up object detection model \cite{jiang2020defense}. The CC3M-QA-DC dataset includes an average of about 30 QA pairs and 12 dense captions per image, covering a wide range of question types (\textit{What}, \textit{How}, \textit{Where}, \textit{Binary}, etc) and a rich corpus of captions. This dataset can be used in combination with the original image-text datasets or the re-generated image-caption datasets (for better performance). Additional examples and detailed information about the dataset can be found in Appendix \ref{section: A}. Our CC3M-QA-DC dataset is publicly available at \href{https://github.com/johncaged/OPT\_Questioner}{https://github.com/johncaged/OPT\_Questioner}.

\section{Experiments}
\label{section: Experiments}

\begin{table*}[htb]
    \caption{Fine-tuning and zero-shot results on 3 common multimodal benchmarks, including COCO text-to-image retrieval, COCO caption and VQA v2. Results on karpathy test split are reported in COCO retrieval and COCO caption and Acc. scores on test-dev are reported in VQA v2. \dag and \ddag denote weight initialized from ImageNet-22K \cite{deng2009imagenet} and CLIP \cite{radford2021learning}, respectively. During pre-training and fine-tuning, all the settings share the same configuration (including training steps) except for pre-training tasks and backbones.}
    \label{tab:fine-tune-zero-shot}
    \begin{tabular}{lcccllllll}
        \toprule
        \multirow{2}{*}{Model} & \multicolumn{3}{c}{Pre-Training Settings} & \multicolumn{3}{c}{COCO Retrieval} & \multicolumn{2}{c}{COCO Caption} & \makecell[c]{VQA v2} \\
        \cmidrule(r){2-4} \cmidrule(r){5-7} \cmidrule(r){8-9} \cmidrule(r){10-10}
        & CC3M & QA & DC & \makecell[c]{R@1} & \makecell[c]{R@5} & \makecell[c]{R@10} & \makecell[c]{CIDEr} & \makecell[c]{SPICE} & \makecell[c]{Acc.} \\
        \midrule
        \multicolumn{10}{l}{\textit{\textbf{Fine-Tuning}}} \\
        \multirow{3}{*}{Swin-B$^{\dag}$ \cite{liu2021swin}-224} & \Checkmark &&& 50.0 & 76.2 & 85.0 & 122.62 & 21.97 & 70.64 \\
        & \Checkmark & \Checkmark && 51.0 (+1.0) & 77.1 (+0.9) & 85.6 (+0.6) & 124.55 (+1.93) & 22.32 (+0.35) & 71.68 (+1.04) \\
        & \Checkmark & \Checkmark & \Checkmark & \textbf{51.7 (+1.7)} & \textbf{77.8 (+1.6)} & \textbf{86.0 (+1.0)} & \textbf{124.97 (+2.35)} & \textbf{22.43 (+0.46)} & \textbf{72.28 (+1.64)} \\
        \midrule
        \multirow{3}{*}{ViT-B/16$^{\dag}$ \cite{dosovitskiy2020image}} & \Checkmark &&& 36.7 & 66.0 & 77.2 & 118.43 & 21.46 & 66.79 \\
        & \Checkmark & \Checkmark && 39.4 (+2.7) & 68.1 (+2.1) & 78.9 (+1.7) & 120.55 (+2.12) & 21.76 (+0.30) & 69.90 (+3.11) \\
        & \Checkmark & \Checkmark & \Checkmark & \textbf{41.1 (+4.4)} & \textbf{69.5 (+3.5)} & \textbf{80.1 (+2.9)} & \textbf{121.16 (+2.73)} & \textbf{22.04 (+0.58)} & \textbf{70.96 (+4.17)} \\
        \midrule
        \multirow{3}{*}{ViT-B/16$^{\ddag}$ \cite{radford2021learning}} & \Checkmark &&& 50.1 & 76.7 & 85.3 & 129.43 & 22.98 & 72.30 \\
        & \Checkmark & \Checkmark && 52.5 (+2.4) & 78.8 (+2.1) & 86.8 (+1.5) & \textbf{130.32 (+0.89)} & \textbf{23.15 (+0.17)} & 74.14 (+1.84) \\
        & \Checkmark & \Checkmark & \Checkmark & \textbf{53.2 (+3.1)} & \textbf{79.2 (+2.5)} & \textbf{87.0 (+1.7)} & 129.49 (+0.06) & 23.02 (+0.04) & \textbf{74.57 (+2.27)} \\
        \midrule
        \multirow{3}{*}{ViT-L/14$^{\ddag}$ \cite{radford2021learning}} & \Checkmark &&& 58.6 & 82.4 & 89.1 & 137.54 & 24.04 & 76.47 \\
        & \Checkmark & \Checkmark && 59.6 (+1.0) & \textbf{83.2 (+0.8)} & \textbf{89.8 (+0.7)} & \textbf{138.09 (+0.55)} & \textbf{24.20 (+0.16)} & 77.81 (+1.34) \\
        & \Checkmark & \Checkmark & \Checkmark & \textbf{59.8 (+1.2)} & \textbf{83.2 (+0.8)} & 89.7 (+0.6) & 137.89 (+0.35) & 24.17 (+0.13) & \textbf{78.30 (+1.83)} \\
        \midrule
        \multicolumn{10}{l}{\textit{\textbf{Zero-Shot}}} \\
        \multirow{3}{*}{Swin-B$^{\dag}$ \cite{liu2021swin}-224} & \Checkmark &&& 36.6 & 62.5 & 72.9 & 44.73 & 11.14 & \makecell[c]{-} \\
        & \Checkmark & \Checkmark && 39.2 (+2.6) & 65.5 (+3.0) & 74.8 (+1.9) & 47.27 (+2.54) & 11.62 (+0.48) & \makecell[c]{44.01}  \\
        & \Checkmark & \Checkmark & \Checkmark & \textbf{41.2 (+4.6)} & \textbf{67.2 (+4.7)} & \textbf{76.4 (+3.5)} & \textbf{49.24 (+4.51)} & \textbf{11.86 (+0.72)} & \makecell[c]{\textbf{45.58}} \\
        \midrule
        \multirow{3}{*}{ViT-B/16$^{\dag}$ \cite{dosovitskiy2020image}} & \Checkmark &&& 22.8 & 47.7 & 60.2 & 40.87 & 10.37 & \makecell[c]{-} \\
        & \Checkmark & \Checkmark && 26.7 (+3.9) & 52.1 (+4.4) & 63.9 (+3.7) & 43.85 (+2.98) & 10.97 (+0.60) & \makecell[c]{41.74} \\
        & \Checkmark & \Checkmark & \Checkmark & \textbf{29.6 (+6.8)} & \textbf{55.3 (+7.6)} & \textbf{66.7 (+6.5)} & \textbf{45.45 (+4.58)} & \textbf{11.16 (+0.79)} & \makecell[c]{\textbf{42.64}} \\
        \midrule
        \multirow{3}{*}{ViT-B/16$^{\ddag}$ \cite{radford2021learning}} & \Checkmark &&& 34.9 & 61.5 & 71.1 & 45.22 & 11.1 & \makecell[c]{-} \\
        & \Checkmark & \Checkmark && 39.4 (+4.5) & 65.9 (+4.4) & 75.7 (+4.6) & 49.32 (+4.1) & 11.71 (+0.61) & \makecell[c]{44.94} \\
        & \Checkmark & \Checkmark & \Checkmark & \textbf{40.9 (+6.0)} & \textbf{66.7 (+5.2)} & \textbf{76.3 (+5.2)} & \textbf{50.84 (+5.62)} & \textbf{12.12 (+1.02)} & \makecell[c]{\textbf{45.65}} \\
        \midrule
        \multirow{3}{*}{ViT-L/14$^{\ddag}$ \cite{radford2021learning}} & \Checkmark &&& 44.2 & 69.2 & 78.2 & 54.73 & 12.37 & \makecell[c]{-} \\
        & \Checkmark & \Checkmark && 48.8 (+4.6) & 72.7 (+3.5) & 80.7 (+2.5) & 56.63 (+1.90) & 12.8 (+0.43) & \makecell[c]{49.92} \\
        & \Checkmark & \Checkmark & \Checkmark & \textbf{49.6 (+5.4)} & \textbf{73.7 (+4.5)} & \textbf{81.2 (+3.0)} & \textbf{58.29 (+3.56)} & \textbf{12.99 (+0.62)} & \makecell[c]{\textbf{50.04}} \\
        \bottomrule
    \end{tabular}
\end{table*}

\subsection{Implementation Details}

\subsubsection{The Generator Model and The Filter Model}

The fundamental model we use for JADE is VALOR$_{L}$ \cite{chen2023valor}. The VALOR model employs ViT-L/14 initialized from CLIP \cite{radford2021learning} as the vision encoder, and BERT$_{base}$ \cite{kenton2019bert} as the text encoder as well as the multimodal decoder. Firstly, it undergoes pre-training on VALOR-1M \cite{chen2023valor}, WebVid-2.5M \cite{bain2021frozen}, CC16M (which includes MSCOCO \cite{linmicrosoft}, Visual Genome \cite{krishna2017visual}, SBU \cite{ordonez2011im2text}, CC3M \cite{sharma2018conceptual} and CC12M \cite{changpinyo2021conceptual}) and HD VILA 10M (randomly sampled from the full HD VILA 100M \cite{xue2022hdvila} dataset), and then it is fine-tuned on VG dataset \cite{krishna2017visual} and a subset of the GQA dataset \cite{hudson2019gqa} (for binary question generation).

The model is implemented using PyTorch and trained on 8 Tesla A100 GPUs. During pre-training, the batch size is 1024 and the learning rate is 1e-4. While during fine-tuning, the batch size and the learning rate are set to 256 and 2e-5, respectively.

\subsubsection{Pre-Training Experiments}

In the pre-training and downstream experiments, we use the same VALOR framework and replace different backbones. Unless otherwise specified, BERT$_{base}$ is used as language backbones. Detailed configuration is introduced in Appendix \ref{section: B}.

\subsection{Pre-Training Experiments}

We conduct experiments using various pre-training tasks, including Image-Text Contrastive Learning (ITC), Image-Text Matching (ITM), Image Captioning (IC), Image-Conditioned Masked Language Modeling (IMLM), Visual Question Answering (VQA), and Image Dense Captioning (DC). The IC, IMLM, VQA, and DC tasks utilize unified Masked Language Modeling, with 60\% of the target tokens randomly masked in IC, IMLM, and DC, and 100\% of the target tokens (i.e., the answer tokens) masked in VQA. The IMLM task uses bi-directional masking, while the IC and DC tasks use causal masking. If not expressly stated, these pre-training tasks share the same image input in one batch and calculate the sum of their loss values as the final loss.


To evaluate our method and dataset, we use three downstream tasks, including Visual Question Answering, Image-Text Retrieval, and Image Captioning. More experiments that are not listed in Section \ref{section: Experiments} can be found in Appendix \ref{section: C}.

\subsubsection{Visual Question Answering (VQA)}

VQA is a commonly used downstream task to evaluate the fine-grained understanding, vision-language alignment, and text generation capabilities of deep learning models. In this section, we choose the VQA v2 benchmark for fine-tuning and evaluation. Other comparative and ablation studies are also mainly based on it, which will be detailed in the following sections.


The fine-tuning results for VQA v2 are shown in Table \ref{tab:fine-tune-zero-shot}, where results on test-dev are reported. Note that all the models show considerable improvement in accuracy when integrating VQA and DC into pre-training, which indicates the effectiveness of our CC3M-QA-DC dataset and multi-task pre-training method. Specifically, the accuracy increases by 1.64\% and 4.17\% in Swin-B$^{\dag}$ and ViT-B/16$^{\dag}$ settings, respectively. Furthermore, it's worth noting that in the ViT-L/14$^{\ddag}$ setting, the accuracy score still shows a significant increase (1.83\%), revealing the potential for improving larger multimodal models using our CC3M-QA-DC dataset.

Zero-shot performance is essential to evaluate the generalization capability of deep learning models in different data domains and tasks. Therefore, we test the zero-shot scores and report the results in Table \ref{tab:fine-tune-zero-shot}. The results show that VQA and DC tasks both help boost zero-shot performance given different backbones, especially in the VQA task. It considerably empowers the model with VQA capabilities without any task-specific learning.

\subsubsection{Image-Text Retrieval}

The image-text retrieval task evaluates the global understanding and matching between images and text. For this task, we use COCO retrieval and report the results for comparison. As shown in Table \ref{tab:fine-tune-zero-shot}, the retrieval scores (R@1, R@5, and R@10) improve when VQA and DC tasks are added, and the R@1 score reaches 59.8 in the ViT-L/14$^{\ddag}$ setting, which is 1.2 higher than the baseline score. Furthermore, our CC3M-QA-DC dataset, along with the original CC3M dataset, can help build better zero-shot models in the image-text retrieval task without access to downstream datasets, as indicated by the zero-shot results in Table \ref{tab:fine-tune-zero-shot}.


\subsubsection{Image Captioning}

The image captioning task measures the generation capability of deep learning models. For evaluation and comparison, we employ the COCO caption benchmark. The fine-tuning results are shown in Table \ref{tab:fine-tune-zero-shot}. The results indicate minor but consistent improvement in SPICE and CIDEr scores. These findings reveal that integrating our VQA and DC tasks into pre-training can improve VQA performance while keeping image captioning performance slightly increasing, at least non-decreasing. This is a significant finding that illuminates the development of stronger and more general foundation models. Moreover, the zero-shot results, displayed in Table \ref{tab:fine-tune-zero-shot}, exhibit clear growth in SPICE and CIDEr scores. This demonstrates a new way of improving multimodal understanding and generation abilities of large pre-trained models.


\subsubsection{Larger Pre-training Data}
\label{section: Larger Pre-training Data}

To further demonstrate the effectiveness of our CC3M-QA-DC dataset and pre-training method, we incorporate a larger dataset, CC12M \cite{changpinyo2021conceptual}, into pre-training and compare it with previous settings. As shown in Table \ref{tab:larger-training}, where VQA v2 Acc. scores on test-dev are reported, even with 5 times the number of image-text pairs, the accuracy in the CC15M (CC3M + CC12M) setting is still 1.61 lower than that in the CC3M + QA + DC setting with the same number of training steps.

In addition, we also simultaneously use both CC15M and CC3M-QA-DC datasets for pre-training and achieve higher accuracy scores. Due to the imbalanced amount of images between the original-caption-related tasks (i.e., ITM, ITC, IMLM, and IC) and our proposed tasks (i.e., VQA and DC), we use a two-pass method in pre-training. In the first pass, image-text pairs are randomly sampled from the CC15M dataset for the original-caption-related tasks, and in the second pass, VQA triplets and dense captions are randomly sampled from our CC3M-QA-DC dataset for our proposed tasks. The final loss for backward is the sum of the loss values from both passes. The results show that our CC3M-QA-DC dataset combined with an even larger image-text dataset can lead to better performance, while generating more VQA and dense captioning data may further improve the performance.

\begin{table}
    \caption{Fine-tuning results using larger pre-training data. ViT-B/16$^{\ddag}$ is used for experiment. All the results adopt the same configuration (including training steps) except for pre-training tasks and data. \dag denotes the QA and dense captioning data is derived from CC3M instead of CC15M.}
    \label{tab:larger-training}
    \begin{tabular}{cccc | l}
        \toprule
        CC3M & QA$^{\dag}$ & DC$^{\dag}$ & CC12M & \makecell[c]{Acc.} \\
        \midrule
        \Checkmark &&&& 72.30 \\
        \Checkmark &&& \Checkmark & 72.96 (+0.66) \\
        \Checkmark & \Checkmark &&& 74.14 (+1.84) \\
        \Checkmark & \Checkmark && \Checkmark & 74.43 (+2.13) \\
        \Checkmark & \Checkmark & \Checkmark && 74.57 (+2.27) \\
        \Checkmark & \Checkmark & \Checkmark & \Checkmark & \textbf{74.88 (+2.58)} \\
        \bottomrule
    \end{tabular}
\end{table}

\subsection{Comparison with State-of-the-Arts}

In this section, we pre-train the VALOR$_{L}$ \cite{chen2023valor} model using the re-captioned CC15M dataset as well as the CC3M-QA-DC dataset, and fine-tune it on the three downstream tasks. It is worth noting that we do not include the MSCOCO, VG, and SBU datasets in the pre-training process to maintain consistency with our previous settings and including these in-domain datasets may lead to further improvements in the results.

\textbf{VQA v2.} As presented in Table \ref{tab:sota-vqa}, VALOR$_{L}$ + QADC achieves a score of 80.12 and 80.16 on the test-dev and test-std benchmarks respectively, outperforming the originally reported performance by a large margin. Our generated CC3M-QA-DC dataset proves to be effective in improving VQA performance and achieving competitive results using a relatively small scale of data. This convincingly demonstrates that applying model-generated QA and dense caption data can boost the model's performance.


\textbf{COCO Retrieval.} As shown in Table \ref{tab:sota-retrieval}, VALOR$_{L}$ + QADC achieves a R@1 score of 64.1, far above the previously reported result involving about 2x pre-training data. Moreover, our method and dataset reach competitive results compared with BLIP \cite{li2022blip}, which adopts 9x pre-training data and similar model scale. This reveals that the model can benefit from fine-grained pre-training tasks and improve its understanding between vision and language modalities.


\textbf{COCO Caption.} To test the generation capability of our method, we fine-tune VALOR$_{L}$ + QADC on the COCO caption benchmark in two ways: cross-entropy fine-tuning and SCST \cite{rennie2017self} fine-tuning. The results are shown in Table \ref{tab:sota-caption}. It can be seen from the table that VALOR$_{L}$ + QADC achieves competitive CIDEr score and SOTA SPICE score, further indicating the effectiveness of our method and dataset.


\begin{table}
    \caption{Performance comparison on VQA v2 benchmark. Results on test-dev and test-std are reported.}
    \label{tab:sota-vqa}
    \begin{tabular}{lll | cc}
        \toprule
        Method & \#Params & \#Data & Test-Dev & Test-Std \\
        \midrule
        \multicolumn{2}{l}{\textit{Closed-ended models}} &&& \\
        FLIP \cite{li2022scaling} & 414M & 400M & 74.70 & - \\
        SimVLM \cite{wang2021simvlm} & 1.4B & 1.8B & 80.03 & 80.34 \\
        Florence \cite{yuan2021florence} & 893M & 900M & 80.16 & 80.36 \\
        OFA$_{large}$ \cite{wang2022ofa} & 472M & 18M & 80.30 & 80.50 \\
        CoCa \cite{yu2022coca} & 2.1B & 4.1B & 82.30 & 82.30 \\
        BEIT-3 \cite{wang2022image} & 1.9B & 21M & \textbf{84.19} & \textbf{84.03} \\
        \midrule
        \multicolumn{2}{l}{\textit{Open-ended generative models}} &&& \\
        ALBEF \cite{li2021align} & 314M & 20M & 75.84 & 76.04 \\
        BLIP \cite{li2022blip} & 385M & 129M & 78.25 & 78.32 \\
        GIT \cite{wang2022git} & 0.7B & 0.8B & 78.60 & 78.80 \\
        VALOR$_{L}$ \cite{chen2023valor} & 486M & 33.5M & 78.46 & 78.62 \\
        VALOR$_{L}$ + QADC & 486M & 14M & \textbf{80.12} & \textbf{80.16} \\
        \bottomrule
    \end{tabular}
\end{table}

\begin{table}
    \caption{Performance comparison on COCO text-to-image retrieval benchmark. R@1, R@5 and R@10 scores on the karpathy test split are reported. \underline{Underline} denotes the second best in the methods.}
    \label{tab:sota-retrieval}
    \begin{tabular}{lll | ccc}
        \toprule
        Method & \#Params & \#Data & R@1 & R@5 & R@10 \\
        \midrule
        ALIGN \cite{jia2021scaling} & 820M & 1.8B & 59.9 & 83.3 & 89.8 \\
        ALBEF \cite{li2021align} & 314M & 20M & 60.7 & 84.3 & 90.5 \\
        FILIP \cite{yaofilip} & 417M & 340M & 61.2 & 84.3 & 90.6 \\
        Florence \cite{yuan2021florence} & 893M & 900M & 63.2 & 85.7 & - \\
        BLIP \cite{li2022blip} & 446M & 129M & \textbf{65.1} & \textbf{86.3} & \textbf{91.8} \\
        VALOR$_{L}$ \cite{chen2023valor} & 486M & 33.5M & 61.4 & 84.4 & 90.9 \\
        VALOR$_{L}$ + QADC & 486M & 14M & \underline{64.1} & \underline{85.4} & \underline{91.2} \\
        \bottomrule
    \end{tabular}
\end{table}

\begin{table}
    \caption{Performance comparison on COCO caption benchmark. CIDEr and SPICE scores on the karpathy test split are reported. \underline{Underline} denotes the second best in the methods. * denotes using SCST \cite{rennie2017self} fine-tuning.}
    \label{tab:sota-caption}
    \begin{tabular}{lll | cc}
        \toprule
        Method & \#Params & \#Data & CIDEr & SPICE \\
        \midrule
        SimVLM \cite{wang2021simvlm} & 1.4B & 1.8B & 143.3 & 25.4 \\
        CoCa \cite{yu2022coca} & 2.1B & 4.1B & 143.6 & 24.7 \\
        LEMON \cite{hu2022scaling} & 675M & 200M & 139.1 & 24.1 \\
        LEMON$^{*}$ \cite{hu2022scaling} & 675M & 200M & 145.5 & 25.5 \\
        GIT$_{L}$$^{*}$ \cite{wang2022git} & 347M & 20M & 144.6 & 25.4 \\
        BLIP \cite{li2022blip} & 446M & 129M & 136.7 & - \\
        Flamingo (80B) \cite{alayrac2022flamingo} & 1.2B & 2.3B & 138.1 & - \\
        VALOR$_{L}$$^{*}$ \cite{chen2023valor} & 486M & 33.5M & \textbf{152.5} & \underline{25.7} \\
        VALOR$_{L}$ + QADC & 486M & 14M & 139.0 & 24.4 \\
        VALOR$_{L}$ + QADC$^{*}$ & 486M & 14M & \underline{147.7} & \textbf{26.7} \\
        \bottomrule
    \end{tabular}
\end{table}

\subsection{Ablation Studies}

\subsubsection{Ablation of Pre-training Tasks}

One crucial assumption about our generated dataset is that, large-scale \textit{silver} data produced by deep learning models are more helpful than small-scale \textit{gold} data annotated mannually. Therefore, in this section, we go deeper into the role of our dataset in the pre-training process. As introduced in the previous sections, the generator and the filter models are derived from fine-tuning on the VG dataset. We replace our CC3M-QA-DC dataset with the original VG-QA dataset and examine the fine-tuning results, which are shown in Table \ref{tab:training-tasks}. The images in the VG dataset and the CC3M dataset differ in sources and amount. Therefore, we apply the two-pass training strategy as described in Section \ref{section: Larger Pre-training Data}. Unless otherwise indicated, ViT-B/16$^{\ddag}$ is used for ablation studies.

It can be seen from the table that, our automatically generated CC3M-QA-DC dataset (e) outperforms the VG dataset (d) even though the VG dataset provides extra image data, which well supports our hypothesis. Through fine-tuning and re-generation, higher scores can be achieved, proving the effectiveness of large-scale model-produced datasets.

Another similar method, proposed by BLIP \cite{li2022blip}, utilizes a pre-trained captioner to produce cleaner caption data, which can also improve the model's performance. Therefore, we follow this method and manage to re-generate CC3M captions, whose result is shown in Table \ref{tab:training-tasks}. The results prove that applying our CC3M-QA-DC dataset (g) can achieve higher accuracy score than the cleaned CC3M dataset (b).

Furthermore, the CC3M-QA setting (c) outperforms the CC3M-only setting (a), which involves 4 training objectives (ITM, ITC, IC and IMLM) and thus occupies more computing resources. It proves that simply applying our CC3M-QA-DC dataset to pre-training for VQA downstream tasks can be a better choice than the original CC3M dataset with caption-related training objectives.

Additionally, we explore the function of the dense captioning data and task. As shown in Table \ref{tab:training-tasks}, integrating the DC task using our CC3M-QA-DC dataset (f) can also improve the VQA performance, mainly due to the involvement of local details and more text corpus.

\begin{table}
    \caption{Ablation of Different Pre-training Tasks. * denotes cleaner caption dataset generated using the method proposed by BLIP \cite{li2022blip}. VQA v2 Acc. scores on test-dev are reported.}
    \label{tab:training-tasks}
    \begin{tabular}{cccccc | l}
        \toprule
        & CC3M & CC3M$^*$ & VG & QA & DC & \makecell[c]{Acc.} \\
        \midrule
        (a) & \Checkmark &&&&& 72.30 \\
        (b) && \Checkmark &&&& 73.86 (+1.56) \\
        (c) &&&& \Checkmark && 73.88 (+1.58) \\
        (d) & \Checkmark && \Checkmark &&& 73.77 (+1.47) \\
        (e) & \Checkmark &&& \Checkmark && 74.14 (+1.84) \\
        (f) & \Checkmark &&&& \Checkmark & 73.64 (+1.34) \\
        (g) & \Checkmark &&& \Checkmark & \Checkmark & 74.57 (+2.27) \\
        (h) && \Checkmark && \Checkmark & \Checkmark & \textbf{75.06 (+2.76)} \\
        \bottomrule
    \end{tabular}
\end{table}

\subsubsection{Ablation of Filtering Strategy}

The filter model in our proposed method has been shown to be effective in removing meaningless or wrongly-answered QA pairs generated by the generator model. In this section, we analyze the performance of the filter model in more detail, and the results are presented in Table \ref{tab:filter}. We observe that applying the filter model significantly improves the quality of the generated QA pairs, as the accuracy scores are much higher compared to the unfiltered setting.

Interestingly, we also observe that considering the answers generated by both the generator and the filter models can lead to a slight increase in performance compared to using only the answers predicted by the filter model. This suggests that the generator model can still provide useful information in certain cases, even if it sometimes generates incorrect or low-quality answers.

Overall, these results demonstrate the importance of using a filter model to improve the quality of generated QA pairs, and suggest that combining the predictions of both the generator and the filter models can lead to even better performance.

\begin{table}
    \caption{Ablation of filtering strategies. VQA v2 Acc. scores on test-dev are reported.}
    \label{tab:filter}
    \begin{tabular}{cc | cc}
        \toprule
        Generator & Filter & Zero-Shot & Fine-Tune \\
        \midrule
        \Checkmark && 44.01 & 73.82 \\
        & \Checkmark & 41.46 & 74.05 \\
        \Checkmark & \Checkmark & \textbf{44.94} & \textbf{74.14} \\
        \bottomrule
    \end{tabular}
\end{table}

\section{Conclusion}

In this paper, we introduce a novel method called Joint QA and DC Generation (JADE) to automatically produce VQA and dense captioning data, by leveraging the ability of large pre-trained multimodal models. Our method considerably decreases the labor cost needed for the creation of VQA and DC datasets. Using JADE, we generate a large-scale VQA and DC dataset derived from the CC3M dataset, named CC3M-QA-DC. Dense experiments show that our proposed method and dataset can greatly improve the models' zero-shot and fine-tuning performance on various understanding and generation tasks, which contributes to the development of generalist and foundation models.

Our work reveals the broad prospects of AI generated data (AIGD) and stronger generalist models without task-specific fine-tuning. We believe that applying larger pre-trained multimodal models and more carefully curated VQA and DC datasets for fine-tuning can generate data of higher quality, thus further improving the effectiveness of our proposed method. In the future, we'll explore VideoQA data generation method and build a Questioner that can generate both VQA and VideoQA datasets, which may equip large pre-trained models with more possible capabilities.

\begin{acks}
This work was supported by the National Key Research and Development Program of China (No. 2020AAA0106400), National Natural Science Foundation of China (U21B2043, 62102419, 62102416).
\end{acks}

\bibliographystyle{ACM-Reference-Format}
\balance
\bibliography{cite}

\appendix
\include{appendix}

\end{sloppypar}
\end{document}

%% file: appendix.tex
\nobalance

\section{Information about CC3M-QA-DC}
\label{section: A}

\subsection{Visualization Examples}

\begin{figure}[h]
  \centering
  \includegraphics[width=\linewidth]{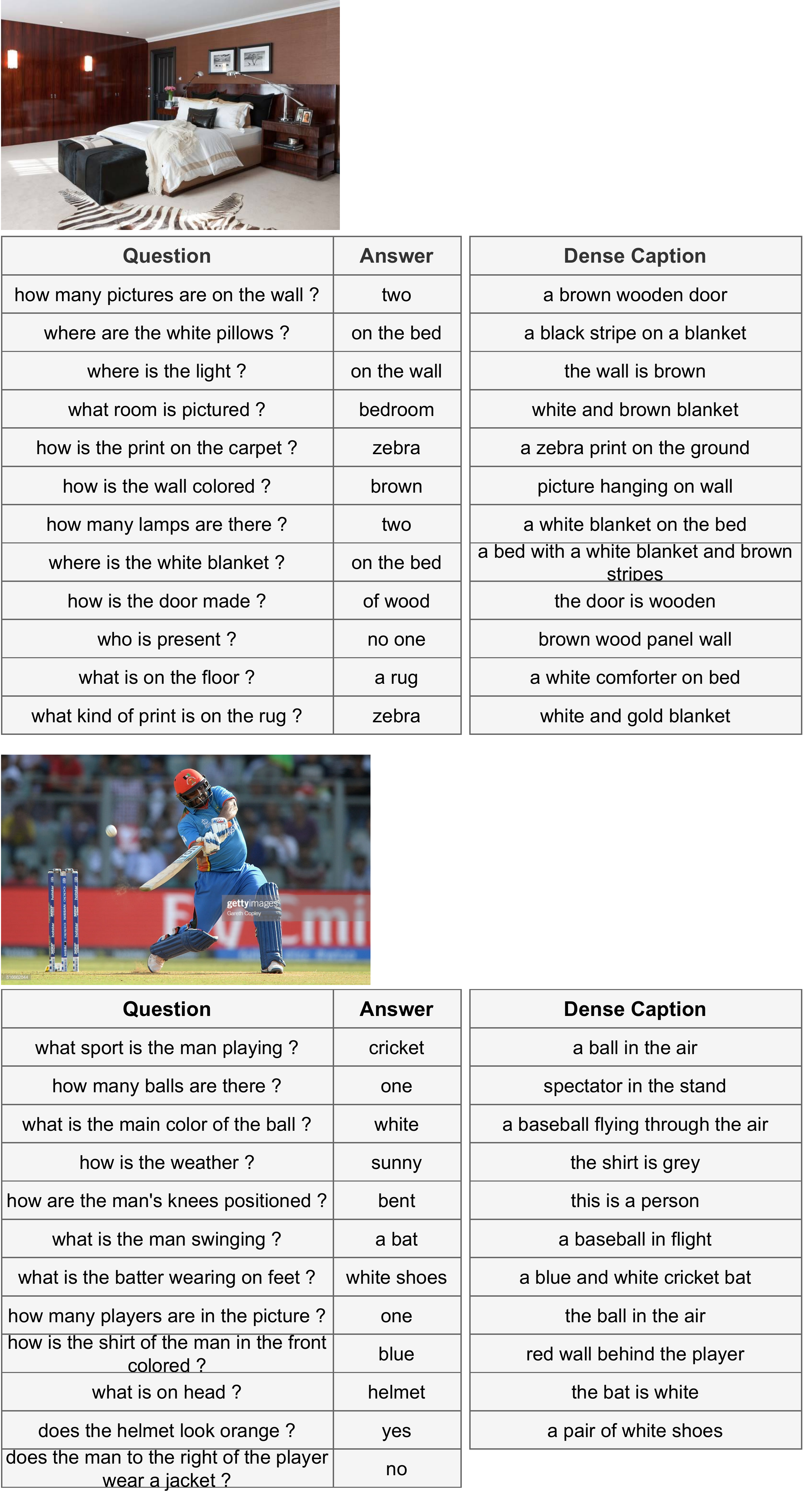}
  \caption{Examples of our generated CC3M-QA-DC dataset. Note that in the generation process, each dense caption is related to multiple QA pairs and after the filtering process, not all the dense captions have the related QA pairs kept. Therefore, in order to show as many examples as possible, the QA pairs and the dense captions are listed independently.}
  \label{fig:example}
\end{figure}

Figure \ref{fig:example} shows some examples of our generated CC3M-QA-DC dataset. It can be seen from the examples that our JADE method can produce detailed questions and dense captions about given images, which is beneficial to multimodal pre-training.

\subsection{Statistics}

In this section, we collect statistics from several datasets, which are displayed in table \ref{tab:datasets}. Compared with other datasets, our CC3M-QA-DC contains more QA and dense caption data, which is more suitable for pre-training.

\begin{table}[h]
    \caption{Statistics of several VQA and DC datasets.}
    \label{tab:datasets}
    \begin{tabular}{lccc}
        \toprule
        Dataset & \#Images & \#QA Pairs & \#Dense Captions \\
        \midrule
        VQA v2 \cite{antol2015vqa} & 123K & 3.7M & - \\
        OK-VQA \cite{okvqa} & 123K & 14K & - \\
        Visual Genome \cite{krishna2017visual} & 108K & 1.7M & 5.4M \\
        CC3M-QA-DC & 2.8M & 89.2M & 37.9M \\
        \bottomrule
    \end{tabular}
\end{table}

\begin{figure}[htb]
  \centering
  \includegraphics[width=0.72\linewidth]{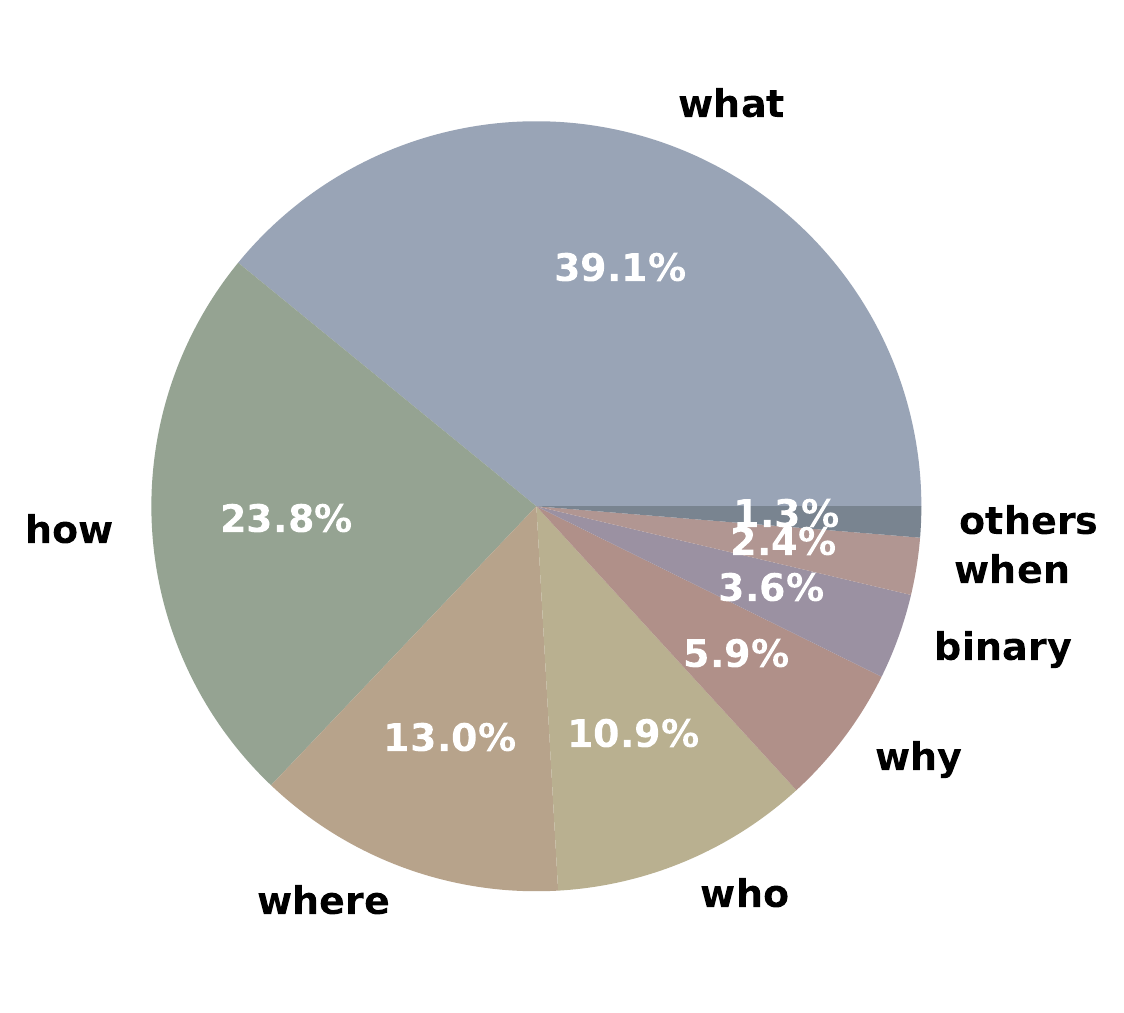}
  \caption{Distribution of question types in the CC3M-QA-DC dataset.}
  \label{fig:pie}
\end{figure}

Furthermore, we count the distribution of question types in our CC3M-QA-DC dataset, which is shown in Figure \ref{fig:pie}.

\section{Experiment Configuration}
\label{section: B}

\subsection{The Generator and The Filter Model}

VALOR$_{L}$ \cite{chen2023valor} is used as the foundation model for JADE. During pre-training, the batch size is 1024 and the learning rate is 1e-4, while during fine-tuning, the batch size and the learning rate are set to 256 and 2e-5, respectively. Unless otherwise specified, all the pre-training process adopts 224 resolution and all the experiments apply AdamW optimizer with $ \beta_{1} = 0.9 $, $ \beta_{2} = 0.98 $ and $ eps = 1e-8 $. Linear warm-up and lr-decay are used.

When fine-tuning the generator and the filter model, we choose batch size, learning rate and resolution as 256, 2e-5 and 224, respectively.

\subsection{Pre-Traing Experiments}

\textbf{Pre-Training}. The pre-training dataset involves CC3M \cite{sharma2018conceptual}, CC12M \cite{changpinyo2021conceptual} and our CC3M-QA-DC. We set the batch size, learning rate, and training epochs to 512, 1e-4, and 10, respectively.

\textbf{COCO Retrieval}. On the COCO retrieval benchmark, we fine-tune the models with a resolution of 384, a batch size of 256, a learning rate of 2e-5, and 50 training epochs. We use the ITM method for testing.

\textbf{COCO Caption}. On the COCO caption benchmark, the models are fine-tuned for 25 epochs with the resolution, batch size and learning rate set to 384, 256 and 2e-5, respectively.

\textbf{VQA v2}. The models are fine-tuned with the resolution, batch size, learning rate, and training epochs set to 392, 256, 2e-5, and 100, respectively.

\subsection{SOTA Experiments}

Most of the configurations in the SOTA experiments are consistent with those in the pre-training experiments, except for the training datasets and VQA v2 training epochs. We adopt re-generated CC12M dataset for pre-training and 200 training epochs in the VQA v2 fine-tuning. In the VQA v2 experiments, the generation process is fully open-ended, without any restrictions or tricks.

\section{More Experiments}
\label{section: C}

\subsection{Ablation of Data Scale}

\begin{figure}[htb]
  \centering
  \includegraphics[width=0.79\linewidth]{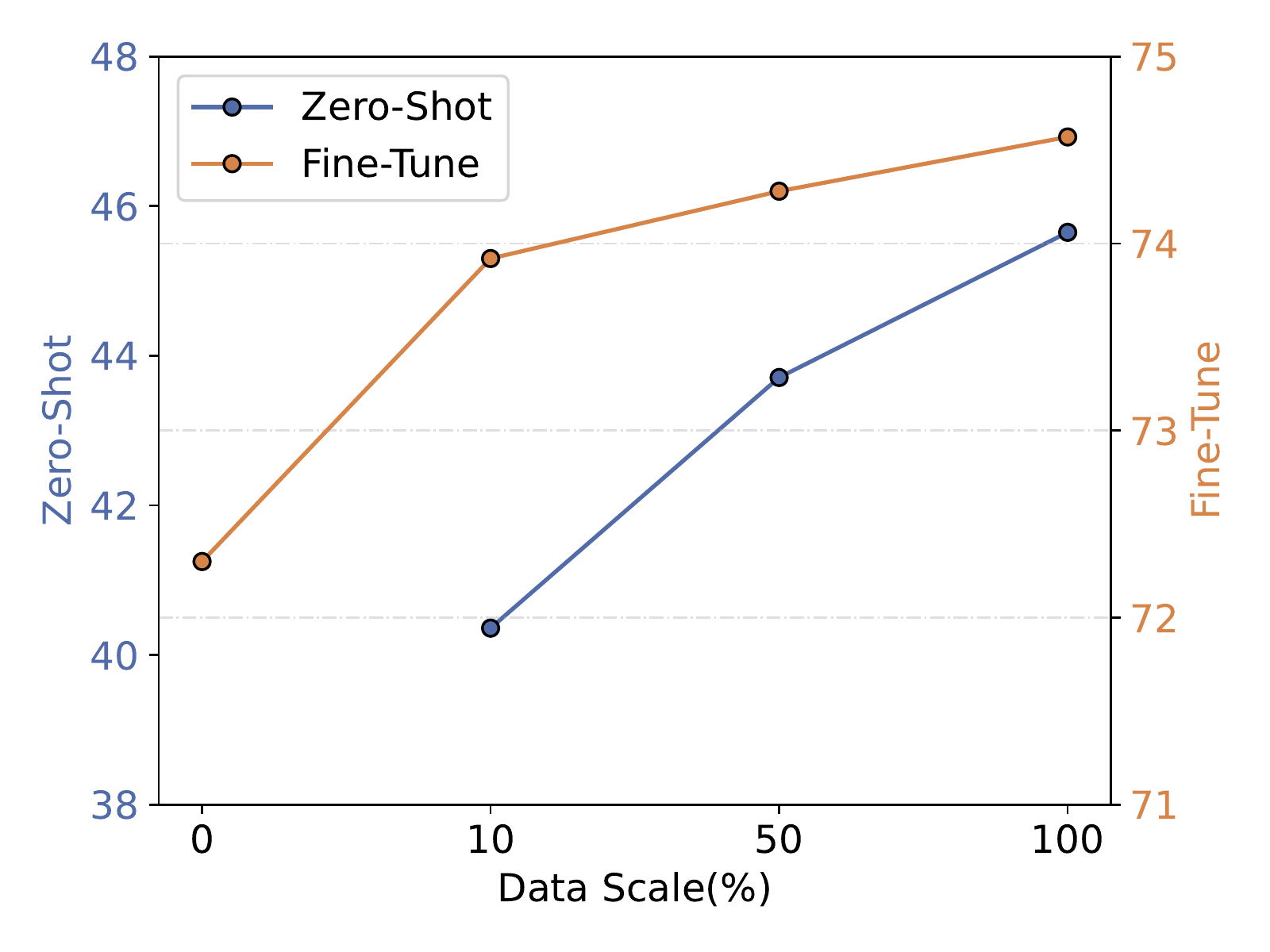}
  \caption{Ablation of different data scales of images in the CC3M-QA-DC dataset. Zero-shot and fine-tuning scores on VQA v2 test-dev are reported.}
  \label{fig:data-scale}
\end{figure}

In this section, we present the results of our experiments where we compare the performance of different data scales by randomly taking subsets of our generated CC3M-QA-DC dataset. As shown in Figure \ref{fig:data-scale}, the accuracy score increases as the scale of the data becomes larger, indicating that more generated VQA and dense captioning data can lead to better performance. These results reveal the potential that generating more VQA and DC data, based on CC12M, LAION-400M \cite{schuhmann2021laion} or even larger datasets, may contribute to the development of generalist and foundation models, which can perform better on zero-shot and fine-tuning tasks.

\subsection{Experiments on More Benchmarks}

To further evaluate the generalization of our proposed method, we carry out experiments on two more downstream benchmarks, i.e., Flickr30K \cite{young2014image} and NLVR2 \cite{suhr2019corpus}.

\subsubsection{Experiments on Flickr30K}

Zero-shot and fine-tuning results of pre-training experiments and comparison with other methods are shown in Table \ref{tab:flickr-pre-training} and Table \ref{tab:flickr-sota}, respectively.

\begin{table}
    \caption{Zero-shot and fine-tuning results on Flickr30K text-to-image retrieval benchmark. ViT-B/16 initialized from CLIP \cite{radford2021learning} is used for experiments.}
    \label{tab:flickr-pre-training}
    \begin{tabular}{llll}
        \toprule
        Method & R@1 & R@5 & R@10 \\
        \midrule
        \multicolumn{4}{l}{\textit{\textbf{Fine-Tuning}}} \\
        CC3M & 76.1 & 93.8 & 96.9 \\
        CC3M + QADC & \textbf{80.4 (+4.3)} & \textbf{95.3 (+1.5)} & \textbf{97.5 (+0.6)} \\
        \midrule
        \multicolumn{4}{l}{\textit{\textbf{Zero-Shot}}} \\
        CC3M & 62.3 & 86.0 & 92.0 \\
        CC3M + QADC & \textbf{69.6 (+7.3)} & \textbf{90.5 (+4.5)} & \textbf{94.6 (+2.6)} \\
        \bottomrule
    \end{tabular}
\end{table}

\begin{table}
    \caption{Zero-shot and fine-tuning performance comparison on Flickr30K text-to-image retrieval benchmark. ViT-L/14 initialized from CLIP \cite{radford2021learning} is used.}
    \label{tab:flickr-sota}
    \begin{tabular}{llll}
        \toprule
        Method & R@1 & R@5 & R@10 \\
        \midrule
        \multicolumn{4}{l}{\textit{\textbf{Fine-Tuning}}} \\
        FILIP \cite{yaofilip} & 87.1 & 97.7 & 99.1 \\
        Florence \cite{yuan2021florence} & 87.9 & 98.1 & - \\
        VALOR$_{L}$ + QADC & \textbf{89.1} & \textbf{98.0} & \textbf{99.3} \\
        \midrule
        \multicolumn{4}{l}{\textit{\textbf{Zero-Shot}}} \\
        Florence \cite{yuan2021florence} & 76.7 & 93.6 & - \\
        Flamingo \cite{alayrac2022flamingo} & 79.5 & 95.3 & 97.9 \\
        CoCa \cite{yu2022coca} & 80.4 & 95.7 & 97.7 \\
        VALOR$_{L}$ + QADC & \textbf{86.1} & \textbf{97.1} & \textbf{98.4} \\
        \bottomrule
    \end{tabular}
\end{table}

\subsubsection{Experiments on NLVR2}

As for the NLVR2 benchmark, given the triplet input, we separately encode the two image-text pairs using VALOR \cite{chen2023valor}, and two output \textit{cls} features are concatenated and then fed into a classification head to predict the label. Fine-tuning results of pre-training experiments and comparison with other methods are shown in Table \ref{tab:nlvr2-pre-training} and Table \ref{tab:nlvr2-sota}, respectively. Accuracy scores on the dev set and the public test set are reported.

\begin{table}
    \caption{Fine-tuning results on NLVR2. ViT-B/16 initialized from CLIP \cite{radford2021learning} is used for experiments.}
    \label{tab:nlvr2-pre-training}
    \begin{tabular}{lll}
        \toprule
        Method & Dev & Test \\
        \midrule
        CC3M & 77.04 & 77.24 \\
        CC3M + QADC & \textbf{80.44 (+3.40)} & \textbf{80.41 (+3.17)} \\
        \bottomrule
    \end{tabular}
\end{table}

\begin{table}
    \caption{Fine-tuning performance comparison on NLVR2. ViT-L/14 initialized from CLIP \cite{radford2021learning} is used.}
    \label{tab:nlvr2-sota}
    \begin{tabular}{lll}
        \toprule
        Method & Dev & Test \\
        \midrule
        BLIP \cite{li2022blip} & 82.2 & 82.2 \\
        SimVLM \cite{wang2021simvlm} & 84.5 & 85.2 \\
        VALOR$_{L}$ + QADC & \textbf{86.1} & \textbf{86.6} \\
        \bottomrule
    \end{tabular}
\end{table}